\documentclass{article}
\usepackage[utf8]{inputenc}
\usepackage[margin = 1in]{geometry}
\usepackage{authblk}

\usepackage{amsmath, amssymb, amsthm, amsfonts, mathtools, multirow, mathrsfs, tabu, booktabs, lipsum, url, subcaption, graphicx,xcolor, tabu,  authblk, hyperref, blindtext}

\usepackage[english]{babel}
\usepackage[shortlabels]{enumitem}
 \usepackage[ruled,vlined, linesnumbered]{algorithm2e}
\usepackage[capbesideposition=outside,capbesidesep=quad]{floatrow}
\usepackage[framemethod=tikz]{mdframed}
\usepackage{bbm}

\allowdisplaybreaks

\newcommand{\appropto}{\mathrel{\vcenter{
  \offinterlineskip\halign{\hfil$##$\cr
    \propto\cr\noalign{\kern2pt}\sim\cr\noalign{\kern-2pt}}}}}
    
\usepackage[backend=biber,style=ieee,sorting=none,natbib=true]{biblatex} 
\usepackage[utf8]{inputenc}
\usepackage[autostyle=true, threshold=2]{csquotes}

\title{\textbf{Classification of Hyperspectral Images Using SVM with Shape-adaptive Reconstruction and Smoothed Total Variation}}
\author[1]{Ruoning Li}
\author[1,2]{Kangning Cui}
\author[1,2]{Raymond H. Chan\footnote{Corresponding Author: raymond.chan@cityu.edu.hk}}
\author[3]{Robert J. Plemmons}
\affil[1]{ Department of Mathematics, City University of Hong Kong}
\affil[2]{ Hong Kong Centre for Cerebro-Cardiovascular Health Engineering}
\affil[3]{ Departments of Mathematics and Computer Science, Wake Forest University}
\date{}                     
\begin{document}
\topmargin=0mm

\maketitle
\begin{abstract}
In this work, a novel algorithm called SVM with Shape-adaptive Reconstruction and Smoothed Total Variation (SaR-SVM-STV) is introduced to classify hyperspectral images, which makes full use of spatial and spectral information. The Shape-adaptive Reconstruction (SaR) is introduced to preprocess each pixel based on the Pearson Correlation between pixels in its shape-adaptive (SA) region. Support Vector Machines (SVMs) are trained to estimate the pixel-wise probability maps of each class. Then the Smoothed Total Variation (STV) model is applied to denoise and generate the final classification map. Experiments show that SaR-SVM-STV outperforms the SVM-STV method with a few training labels, demonstrating the significance of reconstructing hyperspectral images before classification.
\end{abstract}

Smoothed Total Variation, Semi-supervised Learning

\section{Introduction}

Hyperspectral Images (HSIs) often have hundreds of electromagnetic reflectance bands collected by aircraft or satellites that contain discriminative information. HSIs can be represented by $X\in \mathbb{R}^{M\times N\times B}$, where $M$ and $N$ represent the spatial size and $B$ represents the number of spectral bands~\cite{eismann2012hyperspectral}. With the enhancement of the spectral resolution of HSIs,  machine learning approaches can be applied to investigate HSIs in land cover identification, change recognition, and mineral mapping, etc. \cite{chan2020two,cui2021unsupervised,SedaChange2022,8314827}. The ground truth annotations of HSIs are expensive, which motivates us to take less randomly picked labels for training \cite{chan2020two}.

In this work, SVM with Shape-adaptive Reconstruction and Smoothed Total Variation (SaR-SVM-STV) is introduced to classify highly-mixed HSIs in a semi-supervised way. In our previous work, SVM with Smoothed Total Variation (SVM-STV) incorporated a convex variant of the Mumford-Shah denoising algorithm with $\nu$-SVM Classifier to utilize spatial and spectral information \cite{chan2020two}. It has been shown that SVM-STV was highly competitive compared with state-of-the-art algorithms, illustrating the effectiveness of incorporating spatial information into classical pixel-wise algorithms. In this work, spatial information is further utilized by a preprocessing step named Shape-adaptive Reconstruction (SaR). The number of labeled pixels of each class needed to achieve satisfying performance is less than 30.

This paper is organized as follows. Section 2 introduces the related works and the motivation of this work. In Section 3, the workflow of the SaR-SVM-STV Algorithm is introduced. Section 4 shows the numerical results of SaR-SVM-STV compared with $\nu$-SVM and SVM-STV on two benchmark HSI datasets. Section 5 concludes the experiments and discusses the planned future work.

\section{Related Works}
\label{sec:format}

Semi-supervised methods are desirable and computational feasible since they only require a few labeled data for training. The classical semi-supervised pixel-wise algorithms, such as SVMs, performed well for HSIs with small computational cost \cite{melgani2004classification}. A publicly available toolbox called LIBSVM collects the most common SVM models that can be instantly utilized, promoting the application of SVMs \cite{chang2011libsvm}. 

In general, pixel-wise classification methods can be enhanced by analyzing the spatial dependency of nearby pixels, due to the common theoretical-based assumption that the pixels are locally homogeneous in HSIs \cite{liu2013spatial}. In \cite{camps2006composite}, spatial-spectral features are merged to train the SVM Classifier with Mercer's kernels. $K$-means and Principal Component Analysis (PCA) are used in \cite{shu2018learning} to extract features, which are then fed into SVM with RBF kernels. The performance of the above two methods grows significantly compared to pixel-wise classifiers.
The Smoothed Total Variation (STV) model has been used for denoising and segmentation with superb performance~\cite{chan2020two,cai2013two,cai2017three}.
The SVM-STV in \cite{chan2020two} firstly trains a $\nu$-SVM Classifier to estimate the probability maps of each class, and then applies the STV model with $L_1$ and $L_2$ regularizers on the generated 3D probability tensor.

It is customary to employ spatial windows of pixels to extract features in HSIs. Adopting square-shaped fixed-size windows is a straightforward solution. 
A joint sparsity representation (JSR) proposed in \cite{6522858} takes full advantage of the spatial information by constructing JSR on local square patches. However, pixels in a square window, particularly those on the margins or corners, may not have similar spectra as the center pixel.
Nevertheless, fixed-size windows will fail when both spatially homogeneous and heterogeneous regions exist. Nested sliding windows are designed in \cite{ren2021svm} to find the best nested square window that maximizes the mean of the Pearson Correlation between window center and other pixels, and then replace the spectra of center pixels with the product of weighted Pearson Correlation vectors and spectra of pixels in the best nested windows.
In \cite{7276986}, the shape-adaptive (SA) regions of pixels are learned by constructing local smooth irregular polygons in the first principal component.
The local polynomial approximation filtering (LPA) and the intersection of confidence intervals rule (ICI) are applied to determine the size and shape of the polygons \cite{foi2007pointwise}.
Then a shape-adaptive JSR is applied to explore the spatial and spectral information properly within SA regions \cite{7276986}.

In this work, instead of using JSR, we reconstruct the spectrum of each pixel using its SA region.
The main difference between the SaR-SVM-STV proposed in this work and SVM-STV is that the shape-adaptive reconstruction (SaR) is applied to further utilize the spatial features of HSIs before training the $\nu$-SVM. That is, we reconstruct the HSIs by SaR and reduce the dimension of the reconstructed data by PCA, and then perform SVM-STV on the PCA reduced data.
The workflow of the SaR-SVM-STV model is summarized in Algorithm 1 and the following section.

\section{Methodology}
\label{sec:pagestyle}

\begin{algorithm}[t]
\SetAlgoLined
 \KwIn{ HSI, $\Upsilon$ (training set), \\ $\nu$ (tolerance of training error), \\  $\gamma$ (scaling factor of the RBF kernel),  \\ $\beta_1, \beta_2$ (regularization parameters)}
\KwOut{$\hat{Y}$ (classification map)}
Reconstructing the spectrum of each pixel with its Shape-adaptive Regions:
\textbf{$S_c = S_{nbs}\hat{p}(x_c)$}, and get the reconstructed data $\tilde{X}$\;
Reducing the dimension of data by PCA: $X_r=$PCA$(\tilde{X})$\;
Estimating the probability tensor $V$ by using $\nu$-SVMs on the HSI data with parameters $\nu$ and $\gamma$: $V=\nu$-SVMs$(X_r,\nu,\gamma)$\;
Applying STV model on the probability tensor $V$ to get the denoised tensor $U$ for each class:
\textbf{$\min\limits_{{\rm\mathbf{u}}_k} \frac{1}{2}\| {\rm\mathbf{u}}_k-{\rm\mathbf{v}}_k\|_2^2+\beta_1\|{\rm\mathbf{\nabla u}}_k\|_1+\frac{\beta_2}{2}\|{\rm\mathbf{\nabla u}}_k\|_2^2, 
{\rm s.t., }\ {\rm\mathbf{u}}_k|_\Upsilon={\rm\mathbf{v}}_k|_\Upsilon$}\;
The classification map $\hat{Y}$ is obtained by $\hat{Y}_{i,j}=$MaxIndex$(U_{i,j,:})$.

\label{alg: SaR-SVM-STV}
\caption{SVM with Shape-adaptive Reconstruction and Smoothed Total Variation (SaR-SVM-STV)}
\end{algorithm}

\subsection{Shape-adaptive Reconstruction}
The Shape-adaptive Reconstruction (SaR) is introduced as an efficient tool to further utilize spatial features as a preprocessing step. The shape-adaptive (SA) region is generated to pick surrounding pixels that are the closest to the target pixel by the LPA denoising and ICI rule \cite{7276986,foi2007pointwise}. 
The Pearson Correlation coefficients 
${\rm Corr}(x_c,x_i)=\frac{{\rm Cov}(x_c,x_i)}{\sqrt{{\rm Var}(x_c)\cdot {\rm Var}(x_i)}}$ 
are computed to estimate the correlation between the center pixel $x_c$ and the other $n-1$ pixels $x_i$, where $n$ represents the total number of pixels in $x_c$'s SA region. 
Then a vector $p(x_c)\in \mathbb{R}^{n\times 1}$ is formed with the $i$-th entry Corr$(x_c,x_i)$
when $i=1,2,\cdots,n-1$, and the $n$-th entry 1. 
The correlation coefficients are then rescaled as a weight vector $\hat{p}(x_c)=\frac{p(x_c)}{{\rm Sum}(p(x_c))}$, and the spectrum of the center pixel is reconstructed by $S_c = S_{nbs}\hat{p}(x_c)$, where $S_{nbs}\in\mathbb{R}^{B\times n}$ represents the spectra collection of the $n$ pixels in the SA region of pixel $x_c$ and the $n$-th column represents the spectrum of $x_c$. 
Thus, pixels inside SA regions will not contribute to the spectrum of center pixel uniformly, with highly-correlated pixels dominating the spectrum of center pixel as they are associated with large weights in $\hat{p}$. 
PCA dimension reduction is performed to further reduce the noise in HSIs and speed up the training of our proposed model. 
Experiments show that using the data before or after PCA produces similar classification results.

\subsection{SVM with Smoothed Total Variation}
The $\nu$-SVMs are trained by the randomly chosen labels $\Upsilon$ with one-against-one strategy and five-fold cross-validation, and predict the 3D probability map $V\in \mathbb{R}^{M\times N\times K}$. The $k$-th channel $\mathbf{v}_k=V_{:,:,k},k=1,...,K$ represents the probability of the pixels belonging to the class $k$ \cite{chan2020two}. The toolbox of LIBSVM library is used to implement $\nu$-SVM \cite{chang2011libsvm}. 
Then the STV model is adopted to reduce the noise in the probability tensor $V$ by applying the alternating direction method of multipliers (ADMM) with the convergence guaranteed \cite{cai2013two,cai2017three,boyd2011distributed}. The STV model is shown in Step 4 of Algorithm 1. The $\mathbf{u}_k$ represents the smoothed probability map of class $k$. Meanwhile, the labels of pixels in the training set $\Upsilon$ should remain unchanged. The smoothed 3D tensor $U$ is obtained after the STV noise reduction, and the final classification map $\hat{Y}$ is then produced by taking the index of the class that has the maximum probability of each pixel \cite{chan2020two}.

\section{Numerical Results}
\label{sec:typestyle}

\begin{figure}[ht]
    \centering
    \includegraphics[width = 0.6\textwidth]{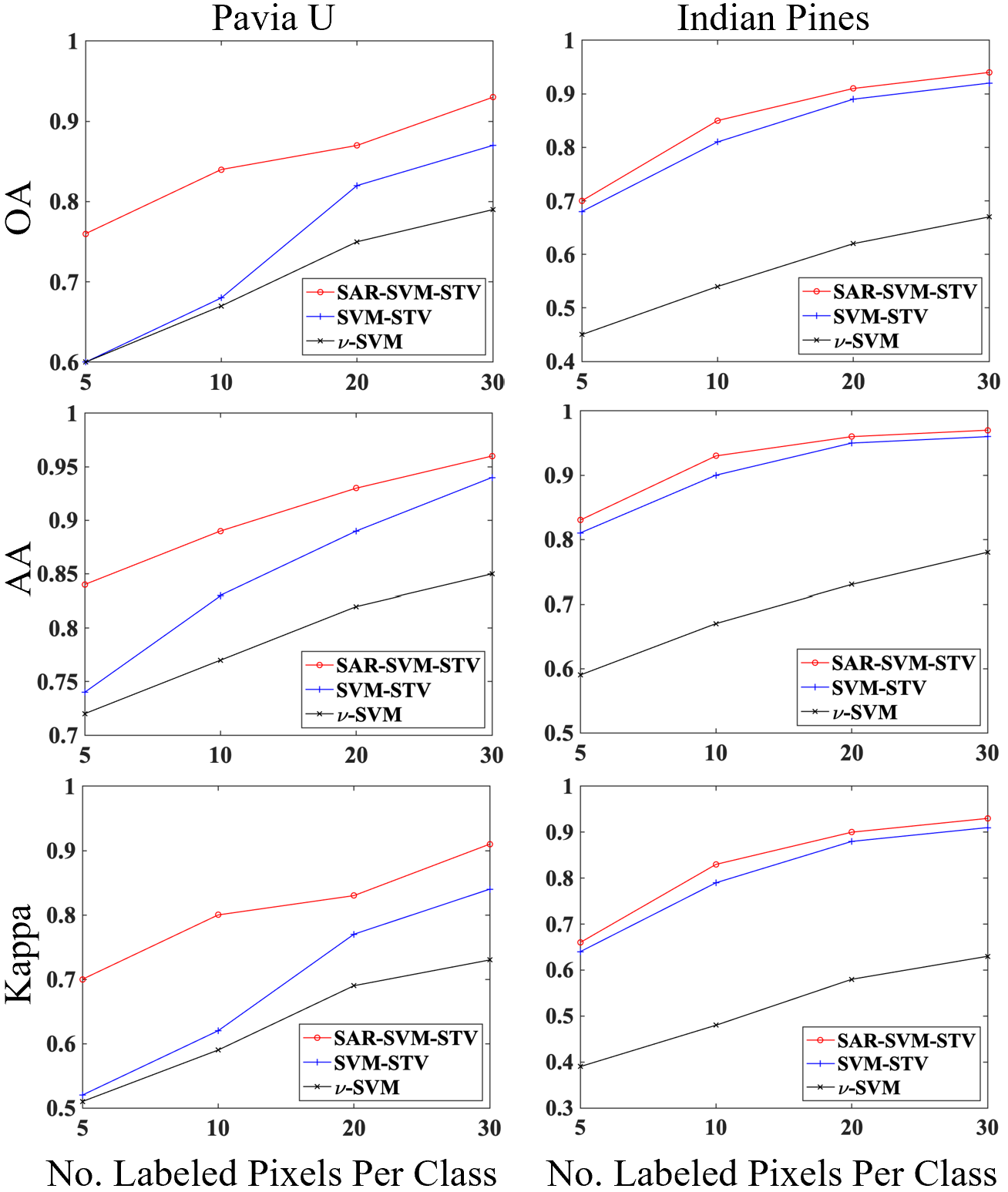}
    \caption{The performance of $\nu$-SVM, SVM-STV, and SaR-SVM-STV on Pavia U and Indian Pines. The two columns present the performance of Pavia U and Indian Pines, and the three rows show the overall accuracy, the average accuracy, and the kappa coefficient of the three algorithms we compare against, respectively.}
    \label{fig:my_label}
\end{figure}

In this section, we show the results of SaR-SVM-STV algorithm compared with $\nu$-SVM and SVM-STV on two publicly available datasets: Indian Pines and University of Pavia (Pavia U). Details of the datasets are summarized in Table 1. 

\begin{table}[htb]
\centering
\caption{Summary of the two HSI datasets}
\label{tab:my-table}
\resizebox{0.7\textwidth}{!}{%
\begin{tabular}{|cccccc|}
\hline
             & Type   & Size                    & No. Classes & Sensor & Location     \\ \hline
Indian Pines & Forest & 145$\times$145$\times$200 & 16          & AVIRIS & Indiana, US  \\
Pavia U      & Urban  & 610$\times$340$\times$103 & 9           & ROSIS  & Pavia, Italy \\ \hline
\end{tabular}%
}
\end{table}

The PCA dimension reduction is used to keep $99.9\%$ of the variance of the reconstructed data. The best $\nu$ and $\gamma$ are selected by the cross-validation of $\nu$-SVM. The best smoothing parameters $\beta_1$ and $\beta_2$ in the STV model are tuned on the probability tensor $V$ that maximize the sum of the overall accuracy (OA), the average accuracy (AA), and the Kappa coefficient in 10 trials. For Pavia U, $\beta_1 = 0.2$, and $\beta_2 = 1$. For Indian Pines, $\beta_1 = 0.2$, and $\beta_2 = 4$.
The source code used in this work can be found at: \href{https://github.com/ckn3/SA-Recon}{github.com/ckn3/SA-Recon}.

Fig. 1 shows the tendency of performance while increasing the labeled pixels randomly picked from each class. The three metrics improve substantially by incorporating more training labels, and the SaR-SVM-STV model outperforms the comparison methods in all cases. 
Fig. 2 shows relationship between the number of labeled pixels involved for training and the count of misclassifications on Indian Pines using SaR-SVM-STV. As the training size increases, the portion of misclassifications reduces monotonically.

Fig. 3 indicates the efficacy of incorporating spatial information. By comparing Figs. 3 (c) and (d), it can be found that although the spatial noise is reduced, the SVM-STV model is misled by misclassified pixels in the predicted map of $\nu$-SVM when only a few labels are used. Applying the SaR and PCA algorithms smooth the HSIs before classification, and thus reduce the number of misclassified pixels in the $\nu$-SVM step. 
The STV model can further tick out the misclassified pixels by enforcing spatial connectivity in probability maps as a postprocessing step, see Fig. 3 (e).

\section{Conclusions}
\label{sec:majhead}

In \cite{chan2020two}, SVM-STV has been shown to outperform many other methods. In this work, we show that adding a preprocessing step, i.e., the Shape-adaptive Reconstruction that reconstruct HSIs using spatial information, can further enhance the performance. The SaR preprocessing and STV postprocessing steps can be easily merged with other semi-supervised or unsupervised algorithms that do not utilize the spatial information of HSIs. The work planned in the near future would be combining the image processing techniques with unsupervised clustering methods~\cite{DVIS,murphy2022multiscale}, and developing the active learning extension where a few carefully-selected annotations are used to train the model~\cite{murphy2018unsupervised,ADVIS}.

\newpage
\begin{figure}[htb]
    \centering
    \includegraphics[width = 0.63\textwidth]{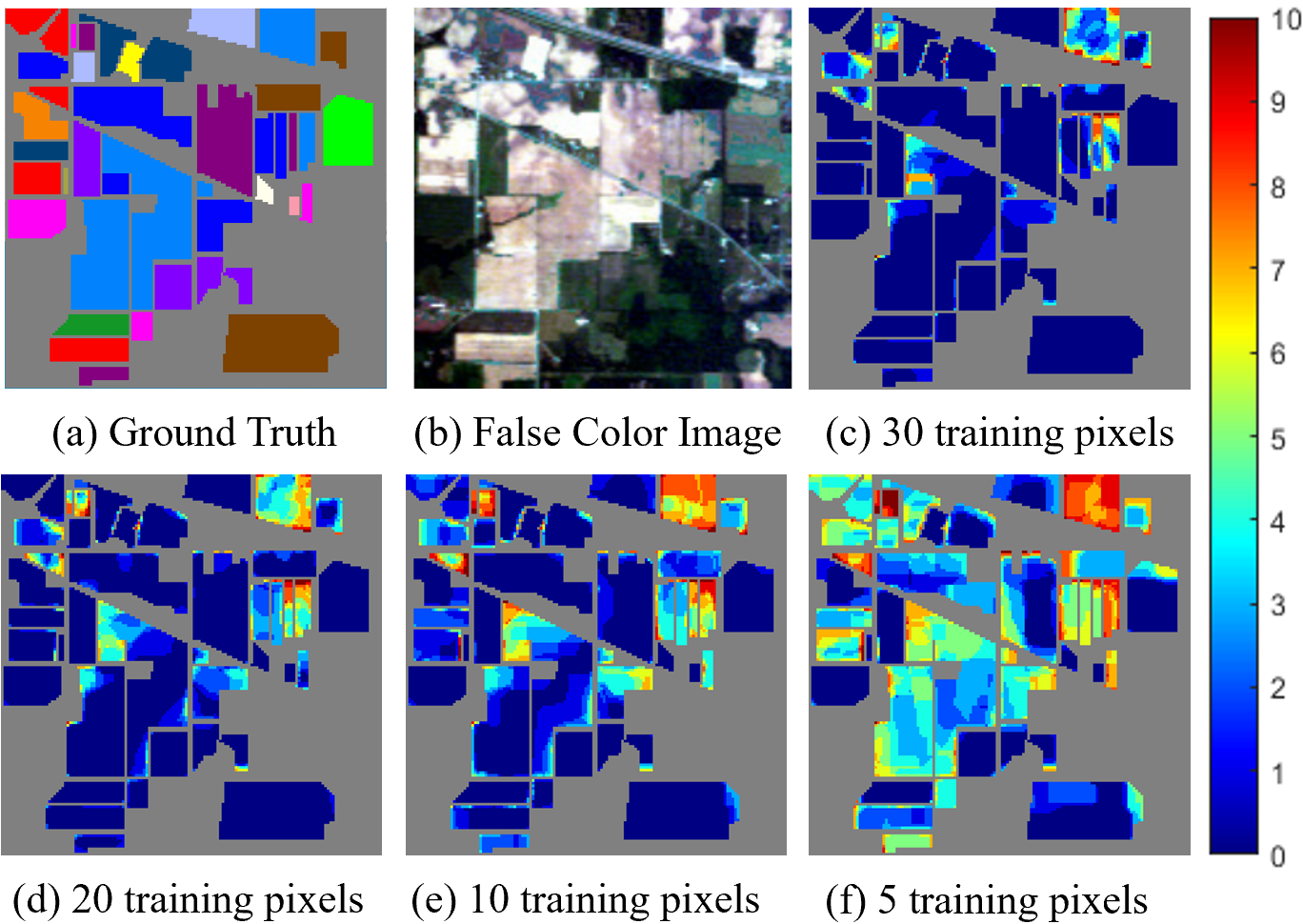}
    \caption{Misclassification heatmaps of Indian Pines using SaR-SVM-STV with different numbers of labeled pixels. Figures (a-b) show the ground truth (each color represents a class) and false-color visualization of Indian Pines. Figures (c-f) show the heatmaps when 30, 20, 10, and 5 labeled pixels per class are used. The colorbar represents the count of misclassifications in 10 trials.}
    \label{fig:ip}
\end{figure}

\begin{figure}[htb]
    \centering
    \includegraphics[width = 0.87\textwidth]{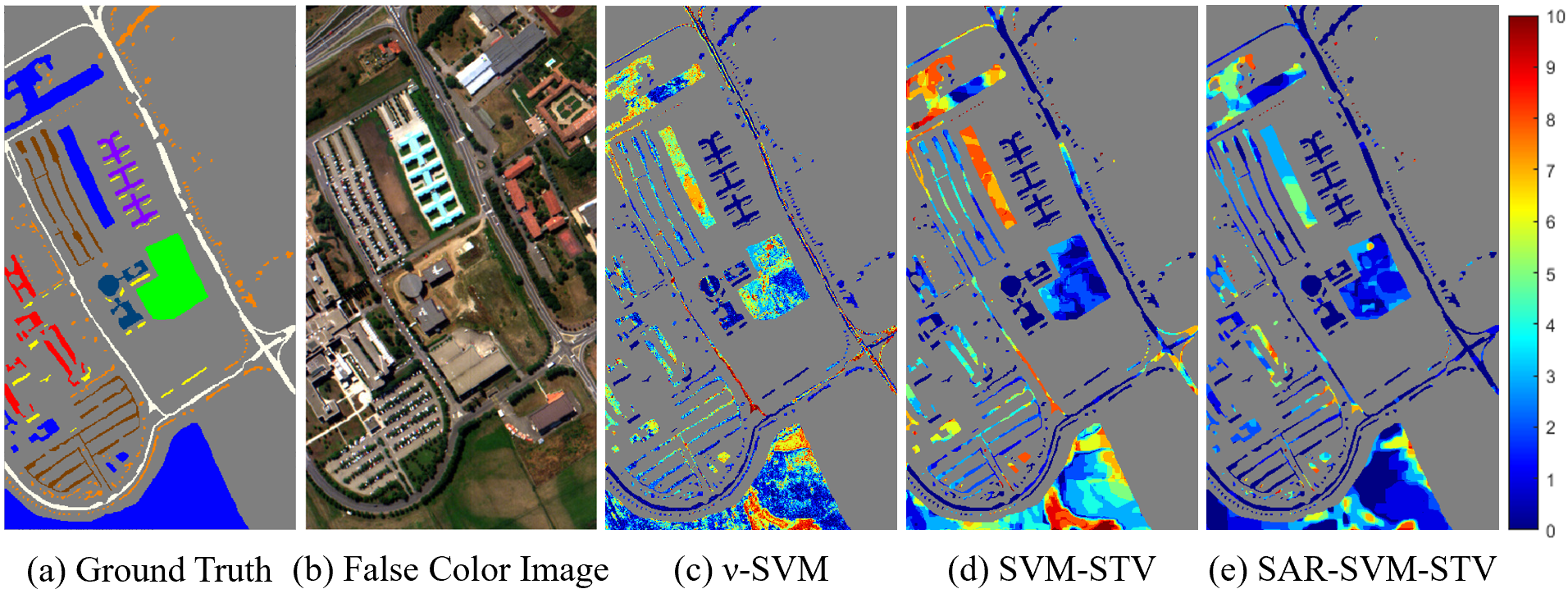}
    \caption{Misclassification heatmaps of Pavia U by comparing the three methods when 10 labeled pixels per class are used for training. Figures (a-b) show the ground truth (each color represents a class) and false-color visualization of Pavia U. Figures (c-e) show the heatmaps of $\nu$-SVM, SVM-STV, and SaR-SVM-STV. The colorbar represents the count of misclassifications in 10 trials.}
    \label{fig:pu}
\end{figure}

\printbibliography 

\end{document}